\documentclass[letterpaper, 10 pt, conference]{ieeeconf}  

\IEEEoverridecommandlockouts                              

\usepackage{graphics} 
\usepackage{epsfig} 
\usepackage{subcaption}
\usepackage{times} 
\usepackage{amsmath} 
\usepackage{amssymb}  
\usepackage{graphicx}
\usepackage{multirow}
\usepackage{amsmath}
\usepackage{array}
\usepackage{bbding}
\usepackage{makecell}
\usepackage[export]{adjustbox}
\usepackage{booktabs}
\usepackage{algorithm}
\usepackage{algpseudocode}

\title{\LARGE \bf
Registering Neural 4D Gaussians for Endoscopic Surgery
}

\author{Yiming Huang, Beilei Cui, Ikemura Kei, Jiekai Zhang, Long Bai, Hongliang Ren$^*$, \textit{Senior Member, IEEE} 
\thanks{This work was supported by Hong Kong RGC GRF 14211420, CRF C4063-18G, NSFC/RGC Joint Research Scheme N\_CUHK420/22; Shenzhen-HK-Macau Technology Research Programme (Type C) STIC Grant 202108233000303; Regional Joint Fund Project 2021B1515120035 (B.02.21.00101) of Guangdong Basic and Applied Research Fund.}
\thanks{$^*$Corresponding author: H. Ren, hlren@ieee.org.}
\thanks{Y. Huang, B. Cui, L. Bai, and H. Ren are with the Department of Electronic Engineering, The Chinese University of Hong Kong (CUHK), Hong Kong, China; and Shenzhen Research Institute, CUHK, Shenzhen, China.
        {(e-mail: yhuangdl@link.cuhk.edu.hk)}}%
\thanks{I. Kei is with KTH, Kungliga Tekniska högskolan Royal Institute of Technology, Sweden.}
\thanks{J. Zhang is with the Hong Kong Applied Science and Technology Research Institute Company Limited, Hong Kong, China.} 
}

\begin{document}

\maketitle
\thispagestyle{empty}
\pagestyle{empty}

\begin{abstract}

The recent advance in neural rendering has enabled the ability to reconstruct high-quality 4D scenes using neural networks. Although 4D neural reconstruction is popular, registration for such representations remains a challenging task, especially for dynamic scene registration in surgical planning and simulation. In this paper, we propose a novel strategy for dynamic surgical neural scene registration. We first utilize 4D Gaussian Splatting to represent the surgical scene and capture both static and dynamic scenes effectively. Then, a spatial aware feature aggregation method, Spatially Weight Cluttering (SWC) is proposed to accurately align the feature between surgical scenes, enabling precise and realistic surgical simulations. Lastly, we present a novel strategy of deformable scene registration to register two dynamic scenes. By incorporating both spatial and temporal information for correspondence matching, our approach achieves superior performance compared to existing registration methods for implicit neural representation. The proposed method has the potential to improve surgical planning and training, ultimately leading to better patient outcomes.
\end{abstract}

\section{Introduction}

Performing pairwise registration of 3D surgical scenes is a vital step for numerous surgical applications, including surgical navigation~\cite{malhotra2023augmented, cui2024surgical} and AR/VR-based surgical reconstructions~\cite{yang2024self}. The process of registering surgical 3D scenes is accompanied by several challenges, including noise, outliers, and partial overlap~\cite{goli2023nerf2nerf}. 3D scene reconstruction has been traditionally achieved with methods that produce explicit 3D representations such as simultaneous localization and mapping (SLAM)~\cite{liu2022sage}, depth-fusion~\cite{recasens2021endo} where the scene is represented with point clouds or meshes. However, registration with explicit representation may perform poorly in surgical scenarios given challenging conditions such as low lighting and lack of texture~\cite{mildenhall2020nerf}. 

\begin{figure}[!th]
\centering
\includegraphics[width=\linewidth]{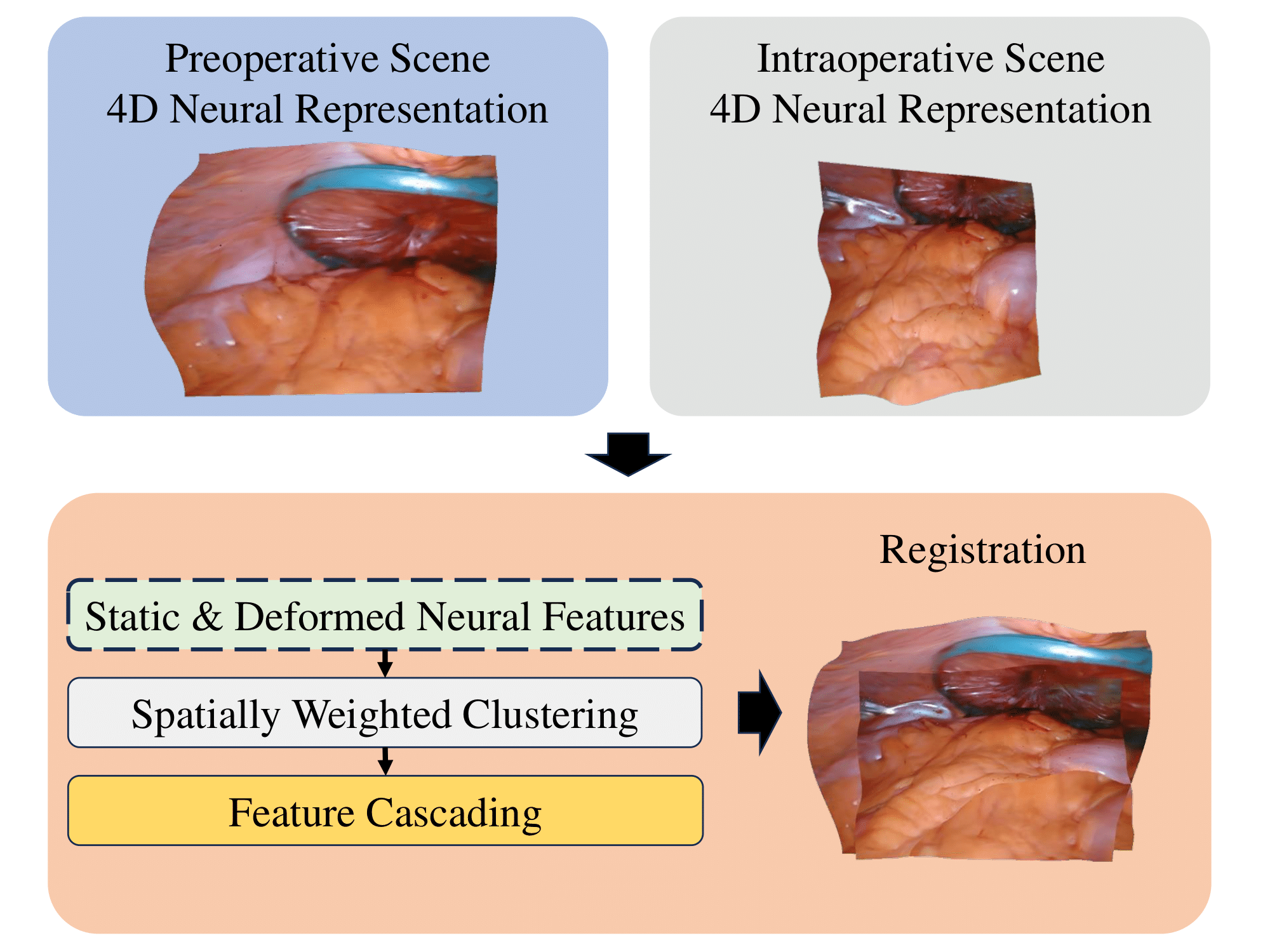}

\caption{\textbf{Overview of the 4D Neural Representation Registration}. With the 2D images of the preoperative scene and intraoperative scene, we train two separate neural networks as implicit 4D scene representations. To establish a full scene for the dynamic surgical reconstruction, we proposed a novel registration method for the implicit 4D representations.}

\label{fig:overview}
\end{figure}

Recently, Neural Radiance Fields (NeRF) techniques have gained significant attention due to their remarkable capability in scene reconstruction~\cite{mildenhall2020nerf, park2021nerfies, park2021hypernerf}. Recent advancements have significantly improved the reconstruction quality of NeRF~\cite{mildenhall2020nerf}. However, previous neural implicit reconstructions mainly targeted the object level~\cite{mildenhall2020nerf} or unbounded scenes within a small field~\cite{mueller2022instant}, of which scenes are static. On the other hand, in dynamic surgical scenes, it is not always possible to capture images with absolute pose information from different viewpoints, such as using endoscope~\cite{zha2023endosurf, wang2022neural}. In such cases, NeRF registration~\cite{dreg} requires synthesizing consistent novel views from multiple NeRFs trained in different coordinate frames, which is incompatible with the deformable surgical scenario.

Since NeRF encodes scenes implicitly using a neural network, instead of explicit representation like point clouds, registration of multiple NeRFs remains a challenging problem that involves non-rigid regression to the implicit space. nerf2nerf~\cite{goli2023nerf2nerf} was the first attempt to address this issue using an extension of the traditional ICP approach~\cite{4767965}, but it requires initializing the correspondence with human-annotation, which can be difficult for real-world surgical scenarios with deformable tissues. On the other hand, DReg-NeRF~\cite{dreg} utilizes a 3D feature pyramid network to register NeRF in voxel grids. 

Despite previous work~\cite{dreg} has achieved satisfactory performance for NeRF registration, it cannot achieve real-time rendering and registration due to its implicit nature. Therefore, we focus on 4D Gaussian Splatting~\cite{Wu_2024_CVPR} for a more explicit and real-time reconstruction of the surgical scene. To this end, we make the effort to explore registering two dynamic neural scenes considering two fundamental questions: 1) How can we adopt registration on 4D Gaussians~\cite{wu20234dgaussians}) in the dynamic complex scenarios in surgeries? 2) How to extract and aggregate features from the implicit encoding of dynamic scenes in 4D Gaussians without any human annotations or initialization?  

We first build up our method based on 4D Gaussian Splatting~\cite{wu20234dgaussians}. Static and dynamic scenes are captured by lightweight MLPs achieving minimum training time and real-time inference speed. We then innovatively propose Spatially Weight Cluttering to automatically extract keypoints based on Gaussian mean and opacity value. We first employ a KMean~\cite{hartigan1979algorithm}-based weighted aggregation to enhance the spatial relationship of neural implicit features within the opacity field. Then we integrate both the static and deformed neural representation by feature cascading to form the initialization of key features for registration. Both the static and deformed Gaussian features are organized into several clusters considering the opacity and spatial information. Finally, we register the output features with~\cite{choi1997performance} and~\cite{4767965}  to output the registration result. The overview of the method is presented in Fig.~\ref{fig:overview}. In contrast to state-of-the-art point cloud registration methods~\cite{9577334, yew2022regtr}, we do not need additional masks for rejecting the outliers of correspondences. Our approach is specifically tailored for dynamic surgical scenes and addresses the unique challenges posed by these scenarios.

In summary, the contributions of our work are:

\begin{itemize}
    \item We first propose Spatially Weight Cluttering (SWC), an efficient method for scene-level implicit 3D feature extraction without any external marking and initializations by humans.
    \item We further develop a novel feature cascading strategy for registering scene-level 4D Gaussian fields on neural-represented surgery with deformable tissues.
    \item Extensive experiments on a publicly available dataset demonstrate the accuracy of our method.
\end{itemize}

To our current understanding, this is the first work of registration for 4D Gaussian representations on a) preoperative and intraoperative Endoscopic scenarios and b) narrow surgical scenes with deformable tissues.

\section{Related Work}

\subsection{Point Cloud Registration.}
The most widely used point cloud registration algorithms in academia and industry are the Iterative Closest Point (ICP)~\cite{4767965} and its derivatives. These methods aim to minimize the distance between two sets of point clouds, but their effectiveness heavily relies on proper initialization and may yield poor results when dealing with partially overlapping point clouds. Another class of algorithms that does not rely on proper initialization and allows only partial overlap of two point clouds is represented by Global Registration~\cite{choi1997performance} and Fast Global Registration (FGR)~\cite{Zhou2016FastGR}. Apart from that, recent research that utilizes the neural network for point-cloud registration, such as~\cite{wang2019deep, wang2019prnet, dang2022learning, jiang2023robust}, is very popular. However, the existing non-learning registration methods cannot align well with the 3D surgical scene due to the loss of the feature from the coarse depth and 2D RGB images, while learning-based methods fail because of the poor quality of the novel views from the implicit scene representation. To deal with these challenges, we proposed a spatial-wise feature aggregation and selection metric, significantly increasing the performance of registration on implicit scene representations.

\subsection{Neural Representation Registration.}
Direct registration of multiple NeRFs is still under-researched. nerf2nerf~\cite{goli2023nerf2nerf} performs NeRF registration via traditional optimization algorithms but requires manual initialization by labeling the correlated points. DReg-NeRF~\cite{dreg} is the first to implement NeRF registration that does not rely on traditional optimization algorithms and does not require manual initialization. Ref-NF~\cite{hausler2024reg} further extends the neural representation registration into unconstrained environments. While both approaches have achieved favorable results at the object level, their performance diminishes when applied to scenes. Moreover, none of these methods support registration in dynamic scenes, which is a crucial consideration within the medical domain where dynamic scenes are prevalent. In contrast to prior work, our study focuses on scene-level registration and offers support for dynamic scenes. While NeRF fails to encode 3D scenes for some specific scenes like surgery, we utilize a 4D dynamic representation~\cite{wu20234dgaussians} as our backbone for testing, achieving superior improvements on both the quantitative and qualitative results.

\subsection{Dynamic Scene Representation.}
Within the surgery domain, the reconstruction of a static 3D scene often falls short of capturing the dynamic nature of the human body. To fully leverage the data from endoscopy and Computed Tomography (CT) examinations, it is essential to employ a dynamic 3D scene representation. Numerous prior studies~\cite{wang2022neural, zha2023endosurf, yang2023neural} have explored the implementation of dynamic surgical scene representations based on Neural Radiance Fields (NeRF). To enable dynamic scene reconstruction, ~\cite{pumarola2020d, wang2022neural} models a deformation network based on a static canonical network. TiNeuVox~\cite{TiNeuVox} adopted a similar approach while accelerating the training speed by introducing optimizable explicit voxel features. However, these approaches lack the ability to represent changes in the scene's topological properties, such as object removal. In contrast, DyNeRF~\cite{li2022neural} employs a set of compact latent codes to capture scene dynamics. Another method, K-planes~\cite{kplanes_2023} decomposes 4D spatiotemporal volumes into six planes followed by a feature decoder for color prediction. Due to its superior performance compared to most dynamic scene representation algorithms and its capacity to represent dynamic scenes through high dimensionality. Recently, inspired by~\cite{kerbl20233d}, Yang~\cite{wu20234dgaussians} proposed another faster scene reconstruction approach based on the Gaussian representation, enabling high-quality reconstruction of the 4D scenes with real-time inference speed.

\begin{figure*}[!ht]
\centering
\includegraphics[width=0.9\textwidth]{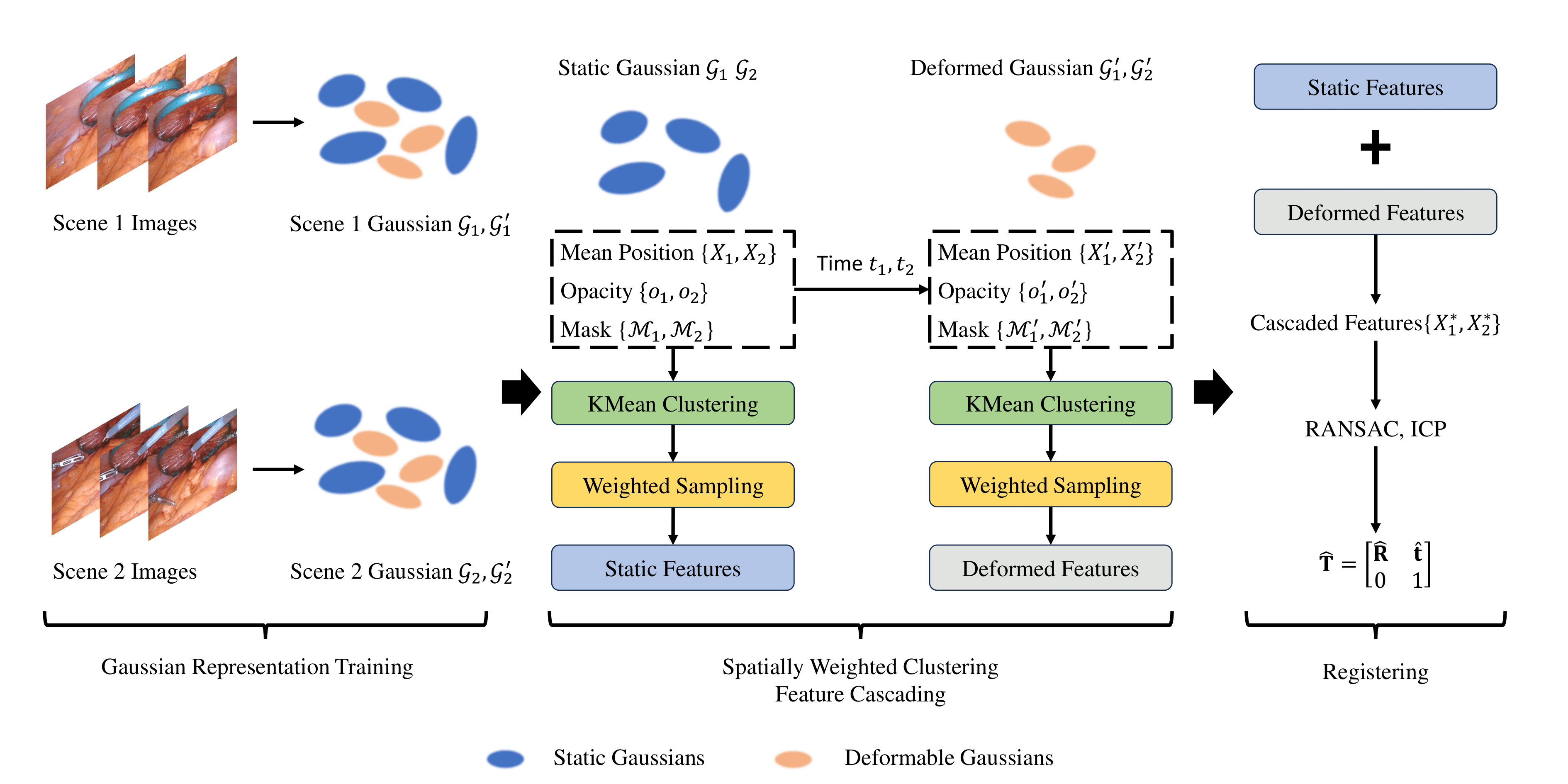}
\caption{\textbf{Illustration of the proposed method}. We first train two 4D neural representations ($\mathcal{G}_1, \mathcal{G}^\prime_1;\mathcal{G}_2, \mathcal{G}^\prime_2$) with images from a preoperative scene 1  and an intraoperative scene 2, then the implicit features $X_1, X_1^\prime; X_2, X_2^\prime$ of 4D Gaussians is passed into our spatially weighted clustering and feature cascading pipeline. Finally, we register the cascaded output $X^*_1, X^*_2$ for the transformation $\mathbf{\hat{T}}$.}
\label{fig:method}
\end{figure*}
\section{Methods}

In this section, we will first elaborate on the preliminary of Gaussian Splatting and then the proposed implicit registration strategy. We demonstrate the proposed spatially weighted clustering method and feature cascading for registration between multi-scene neural representations. The overview of the method is shown in Fig.~\ref{fig:method}.

\subsection{Gaussian Splatting} Kerbl \textsl{et al.}~\cite{kerbl20233d} proposed 3D Gaussian Splatting which learn a 3D differentiable Gaussian representation for rapid scene reconstruction. By definition, the 3D Gaussian are representing by a covariance matrix ${\rm\mathbf{\Sigma}} = \mathbf{R}\mathbf{S}\mathbf{S}^T\mathbf{R}^T$ and a center point $x$ as the following:

\begin{equation}
\label{formula:gaussian's formula}
    G(x)=e^{-\frac{1}{2}(x-\mu)^T{\rm \mathbf{\Sigma}}^{-1}(x-\mu)},
\end{equation}

\noindent where  $\mathbf{S}$ and $\mathbf{R}$ are scaling rotation. Following~\cite{yifan2019differentiablesplatting}, with the viewing transform $\mathbf{W}$ and the Jacobian of the affine approximation of the projective transformation $\mathbf{J}$, the covariance can be further projected into camera space as ${\rm\mathbf{\Sigma}}^{\prime} = \mathbf{J}\mathbf{W}{\rm \mathbf{\Sigma}} \mathbf{W}^T\mathbf{J}^T$. The final rendering is described as:

\begin{equation}
     \hat{C} = \sum_{i\in N}c_i \alpha_i \prod_{j=1}^{i-1} (1-\alpha_i),
\end{equation}
\noindent where $\hat{C}$ is the final color integration for $K$ points. $c_i$ is the color from the spherical harmonics and $\alpha_i$ is the density from the 2D covariance $\mathbf{\Sigma}^\prime$ and the learned opacity $o$.

\label{subsec:4D Gaussian Splatting}
Followed~\cite{wu20234dgaussians}, we further employ deformation fields $\mathcal{F}$ to predict each 3D Gaussian's deformation at a given timestamp $t$, and represent the deformable Gaussian as:




\begin{equation}
\begin{aligned}
    \mathcal{G}^\prime=\Delta \mathcal{G} + \mathcal{G}, \Delta \mathcal{G}=\mathcal{F}(\mathcal{G}, t)
\end{aligned}
\end{equation}

\noindent where the static 3D Gaussian $\mathcal{G}$ and its deformation $\Delta \mathcal{G}$ is combined to represent the scene deformation. To encode spatial-temporal features, a encoder $\mathcal{H}$ is defined with multi-resolution Hexplanes $R_l(i,j)$ with a tiny MLP $\phi_d$,  $\mathcal{H}(\mathcal{G},t)=\{R_l(i,j), \phi_d | (i,j) \in \{(x,y),(x,z),(y,z), (x,t),(y,t),(z,t)\}, l \in \{1,2\}\}$, The encoded feature is described as $f_d=\mathcal{H}(\mathcal{G}, t)$.

To decode the deformation, a multi-head Gaussian deformation decoder $\mathcal{D}=\{\phi_\mu, \phi_r, \phi_s, \phi_o, \phi_\mathbf{SH}\}$ is proposed, which includes the deformation of position, rotation, scaling, opacity and spherical harmonics $\mathbf{SH}$ with five tiny MLPs. The final definition of 4D Gaussian can be expressed as:
\begin{equation}
    \begin{aligned}
        \mathcal{G}^\prime = \{&\mu+\phi_\mu(f_d), r+\phi_r(f_d), s+\phi_s(f_d),\\ 
        &o+\phi_o(f_d), \mathbf{SH}+\phi_{\mathbf{SH}}(f_d)\}\\
        =\{&\mu+\Delta\mu, \mathbf{R}+\Delta \mathbf{R}, \mathbf{S}+\Delta \mathbf{S},\\
        &o+\Delta o, \mathbf{SH}+\Delta \mathbf{SH}\}
    \end{aligned}
\end{equation}
Following~\cite{huang2024endo4dgs}, we also adopt the pre-trained depth estimation model~\cite{depthanything} for more robust deformable surgical scene reconstruction in terms of geometry. 



\subsection{Spatially Weighted Clustering}
To remove redundant spatial features with low opacity, and achieve a memory and time-efficient registration, we proposed a Spatially Weighted Clustering (SWC) algorithm. Given a pre-trained Gaussian representation $\mathcal{G}$, we utilize the mean position $x\in X$ and the opacity $o$ to perform the spatial feature extraction. We adopt a threshold $\epsilon$ to filter the principal Guassians from the trained Gaussian representation and obtain an opacity mask $\mathcal{M} = o \leq \epsilon$. 

Next, with the opacity mask $\mathcal{M}$, the spatial position set $X$, and the opacity $\sigma$, we further implement the Spatially Weighted Clustering algorithm for Gaussian feature aggregation and selection. 

\begin{algorithm}
\renewcommand{\algorithmicrequire}{\textbf{Input:}}
\renewcommand{\algorithmicensure}{\textbf{Output:}}
\caption{Spatially Weighted Clustering}\label{alg:cluster}
\begin{algorithmic}
    \Require 
    \State Opatity $o$
    \State Opacity mask $\mathcal{M}$
    \State Mean positions $X$
    \State Number of clusters $N$
    \State Drop rate $\delta$
    \Ensure 
    \State Weighted clustered mean positions $X^*$
    \\
    \State $X^* \gets X[\mathcal{M}]$ \hfill{// Initializing}
    \State $o^* \gets o[\mathcal{M}]$ 
    \State $Bins \gets \text{KMean}(X^*, N)$ \hfill{// Clustering}
    \\
    \For{$i\leftarrow 1$ to $N$}{
        \State $n_i \gets \text{size}(Bins(i)) \times (1-\delta)$ \hfill{// Weighting}
        \State $o^*_i \gets o^*[Bins(i)]$
        \State $\mathcal{M}_i \gets \text{topK}(o^*_i, n_i)$
        \State $\mathcal{M^*} \gets \mathcal{M^*} \odot \mathcal{M}_i$
    \EndFor}\\
    return $X^* \gets X^*[\mathcal{M^*}]$
\end{algorithmic}
\end{algorithm}

The pseudo-code of our proposed SWC is presented in Algorithm~\ref{alg:cluster}. The algorithm is initialized with the opacity filtered mean $X^*$ , opacity $o^*$ and $N$ number of clusters N for $\text{KMean}$~\cite{hartigan1979algorithm} method. For each iteration, the opacity $o^*$ drops the last $\delta$ percentage with a top $K$ elements selection function $\text{topK}$, weighted by the size of the cluster $\text{size}(Bins(i))$. After updating the mask $\mathcal{M}$ in each iteration, the optimal $X^*$ is obtained.   

\subsection{Feature Cascading for Deformable Scene Registration} We define our problem as registering two scenes $\mathcal{G}_1, \mathcal{G}_2$ with Gaussian representation at time $t_1, t_2$ respectively. The proposed registration method only utilizes the two individually trained Gaussian implicit field models as Inputs. By utilizing the trained network $\mathcal{F}_1, \mathcal{F}_2$, we calculate the deformed Gaussians as $\mathcal{G}^\prime_1 = \mathcal{G}_1+\mathcal{F}(\mathcal{G}_1, t_1), \mathcal{G}^\prime_2 = \mathcal{G}_2+\mathcal{F}(\mathcal{G}_2, t_2)$. With the mean positions $X_1\in \mathcal{G}_1, X_2\in \mathcal{G}_2$, opacities $o_1\in \mathcal{G}_1, o_2\in \mathcal{G}_2$, and their deformed representation $X_1^\prime\in \mathcal{G}^\prime_1, X_2^\prime\in \mathcal{G}^\prime_2, o_1^\prime\in \mathcal{G}^\prime_1, o_2^\prime\in \mathcal{G}^\prime_2$, we obtain the masks $\mathcal{M}_1, \mathcal{M}_2, \mathcal{M}^\prime_1, \mathcal{M}^\prime_2$. To enable a more robust registration in the deformable surgical scene, we adopt both the static and deformed Guassians for feature extraction. We apply the SWC in a cascading way to extract the optimal neural representation $X^*_1, X^*_2$ for registration as follows:
\begin{align}
    X^*_1 &= \text{SWC}([X_1, X_1^\prime], [o_1, o_1^\prime], [\mathcal{M}_1, \mathcal{M}_1^\prime], N, \delta)\\
    X^*_2 &= \text{SWC}([X_2, X_2^\prime], [o_2, o_2^\prime], [\mathcal{M}_2, \mathcal{M}_2^\prime], N, \delta),
\end{align}

\noindent where, $\text{SWC}(\cdot)$ is the Spatially Weighted Clustering, and $[\cdot]$ indicates concatenation. With $X^*_1, X^*_2$ as input, we estimate the relative 6-DoF transformation $\mathbf{\hat{T}}$ by utilizing the RANSAC~\cite{choi1997performance} and ICP~\cite{4767965}:
\begin{equation}
    \mathbf{\hat{T}} = \text{ICP}(\text{RANSAC}(X^*_1, X^*_2))=\left[\begin{array}{cc}
   \mathbf{\hat{R}}  & \mathbf{\hat{t}} \\
    0 & 1
\end{array}\right] ,
\end{equation}

\noindent where $\mathbf{\hat{R}}, \mathbf{\hat{t}}$ are the rotation and translation from the coordinate of scene 1 to scene 2, respectively. With the cascading integration of the static and deformed Guassian features, we can not only capture the static anatomical structures but also the motion, distortion, and shape variations of tissues, organs, and surgical instruments, achieving a more comprehensive and accurate registration between deformable surgical scenes.

\section{Experiments}

\begin{table*}[!t]\centering
\footnotesize
\resizebox{0.85\textwidth}{!}{
\begin{tabular}{lccccccc}\toprule
&DReg-NeRF~\cite{dreg} & RANSAC~\cite{choi1997performance} &RANSAC~\cite{choi1997performance} + ICP~\cite{4767965} & FGR~\cite{fgr} &FGR~\cite{fgr} + ICP~\cite{4767965}  & Ours \\\midrule
$\Delta R$ average$\downarrow$ &179.86 & 53.75 & 53.36 & 55.42 & 55.14 & \textbf{33.78} \\
$\Delta t$ average (mm)$\downarrow$&158.54 & 22.87 & 25.89 & 36.96 & 34.93 & \textbf{5.08}\\
Time (s)$\downarrow$ & $1.00^*$ & 28.37 & 28.91 & 212.81 & 213.37 & \textbf{0.56}\\
\bottomrule
\end{tabular}}
\caption{\textbf{Quantitative results of the camera trajectories on StereoMIS~\cite{hayoz2023pose}.} Our method achieves the best result on the rotation and translation. All methods are applied to two pre-trained scenes described by 4D Gaussians~\cite{wu20234dgaussians} except DReg-NeRF~\cite{dreg}. $*$ denotes the registration network of DReg-NeRF~\cite{dreg} is trained on the endoscopic data for more than 24 hours. The highlighted result is the best.}
\label{tab:all_results}
\end{table*}

This subsection first introduces the dataset used for the experiments, then the experiment setup and evaluation methodology, and finally, a comparison of the performance of our method with that of DReg-NeRF\cite{dreg}, RANSAC~\cite{choi1997performance} and FGR\cite{fgr} on the dataset.
\subsection{Datasets}
To demonstrate the performance of scene-level dynamic scene registration in surgery scenarios, StereoMIS~\cite{hayoz2023pose} was chosen as the primary test dataset. The dataset is for SLAM in endoscopic surgery and was recorded by da Vinci Xi surgical robot and contains a total of 16 video clips, two of which will be selected for experimentation and comparison. The selected scenarios will have many challenging elements, including breathing motion, tissue deformation, resections, bleeding, and surgical tools. The objective of using this dataset is to demonstrate that the proposed algorithm is applicable to surgical scenarios and is capable of handling dynamic scenarios. Here, we select to operate the experiments on the first training scene of the StereoMIS~\cite{hayoz2023pose} dataset, where we sample 200 images from the $801 - 1000$ frames for the preoperative scene reconstruction, and $13901 - 14100$ frames for the intraoperative scene reconstruction. Each split of the training data contains $200$ images that are $2\times$ downsample from $1280 \times 1024$ to $640 \times 512$, respectively.

\subsection{Baselines}
To evaluate the performance of our strategy, we adopt two baselines from the others, which are InstantNGP~\cite{mueller2022instant}, and 4D Gaussian Splatting~\cite{wu20234dgaussians}. We config the occupation grid in InstantNGP~\cite{mueller2022instant} with a resolution of $128\times 128\times 128$, and the grid resolution of the 4D Gaussian as $64 \times 64 \times 64$. While both baselines are re-implemented with the open-source code except for the depth rendering and supervision, we compare the results of both baselines equipping with our methods.

\subsection{Experiment Setup}
In the NeRF registration part of~\cite{dreg}, the number of clustered groups is $N=5$, and we use an opacity threshold $\epsilon=0.8$ for opacity-based filtering, and a drop rate $\delta=0.5$ for clustering. Following the setting in DReg-NeRF~\cite{dreg}, we train two separated InstantNGPs~\cite{mueller2022instant} and the registration network with 10000 iterations. The training of InstantNGP~\cite{mueller2022instant} and the registration network in DReg-NeRF~\cite{dreg} takes 3 hours and 24 hours, respectively. The training of the 4D Gaussians with the same setting as~\cite{wu20234dgaussians} takes 8 minutes. All experiments were conducted on a single RTX 4090 GPU. To evaluate the registration between two scenes, we adopt a pre-generated random transformation matrix that transfers all poses from the scene 1 coordinate to the scene 2 coordinate, and we compare the estimated transformation with the pre-generated matrix which serves as the ground truth.

\subsection{Evaluation}
Our algorithm is compared with the state-of-the-art learning-based algorithm DReg-NeRF~\cite{dreg} (trained on the endoscopic data within the same dataset) and the non-learning-based algorithm RANSAC (RAndom SAmple Consensus)~\cite{choi1997performance}, FGR (Fast Global Registration)~\cite{Zhou2016FastGR} with the choice of ICP (Iterative Closest Point)~\cite{4767965} refinement. To evaluate each method, we utilize the point cloud-like neural feature as input. Similar to~\cite{dreg}, the mean of the rotation ($\Delta R$ average) and translation($\Delta t$ average) errors of all the algorithms are computed for comparison, both being as low as possible. Additionally, We also evaluate and compare the registration time (Time (s)).

\subsection{Result}

\subsubsection{Quantitative results}
The following Table~\ref{tab:all_results} shows the error mean for the proposed algorithms and other approaches. Compared with the other approaches, with the application of our method, we achieve the best result in both the average of rotation and translation, outperforming the traditional methods such as RANSAC~\cite{choi1997performance} and FGR~\cite{Zhou2016FastGR} and the SOTA deep learning-based method DReg-NeRF~\cite{dreg}. While the original DReg-NeRF method fails to register the implicit representation in both the rotation and translation ($\Delta R$ average of 179.86 and $\Delta t$ average of 158.54), with the application of our proposed spatially weighted clustering and feature cascading strategy, the result is significantly improved and competes the best ($\Delta R$ average of 33.78 and $\Delta t$ average of 5.08) against all other baselines. This result illustrates that with our approach, we can extract the key features more precisely and efficiently.


In addition, the training of the deep learning-based registration network in DReg-NeRF takes hours, while the registration of our proposed method on 4D Gaussian only takes less than one second. In terms of the registration time, with our proposed SWC strategy, we achieve the shortest time of $0.56$ second. We argue that our solution on 4D Gaussian is meaningful for real-world applications such as real-time SLAM in endoscope surgery and visual serving of surgical robots. 


\begin{table}[!t]\centering
\footnotesize
\resizebox{\linewidth}{!}{
\begin{tabular}{lccccccc}\toprule
& $N$=0 & $N$=5 & $N$=10 & $N$=15 & $N$=20 & $N$=25 \\\midrule
$\Delta R$ average$\downarrow$ &42.05 & \textbf{33.78} & 36.08 & 96.05 & 140.18 & 125.71 \\
$\Delta t$ average (mm)$\downarrow$ &22.73 & \textbf{5.08} & 12.04 & 13.62 & 17.04 & 25.68 \\
\bottomrule
\end{tabular}}
\caption{\textbf{Ablation experiment of the clustering number.} To evaluate the influence of the number of clusters, we adjust the number $N$ by a step size of 5. $N=0$ indicates that the spatially weighted clustering is removed.}
\label{tab:cluster}
\end{table}


\begin{table}[!t]\centering
\footnotesize
\resizebox{0.85\linewidth}{!}{
\begin{tabular}{cc|ccc}\toprule
\thead{Static\\Features } &\thead{Deformed\\Features }& \thead{$\Delta R$ average$\downarrow$} & \thead{$\Delta t$ average (mm)$\downarrow$}   \\ \midrule
\Checkmark & \XSolidBrush &49.48 & 6.40   \\
 \XSolidBrush &\Checkmark &88.36 & 14.48  \\
\Checkmark & \Checkmark &\textbf{33.78}  & \textbf{5.08}  \\
\bottomrule
\end{tabular}}
\caption{\textbf{Ablation experiment of the cascading strategy.} We analyze the contribution of components in the cascading features by removing (i) the static features, and (ii) the deformed features. The best results are in bold.}
\label{tab:cascad}
\end{table}

\subsubsection{Ablation studies}
To further evaluate the contributions of each component in our design, we conducted two ablation studies on the StereoMIS dataset. The first ablation experiment shown in Table~\ref{tab:cluster} focused on the clustering number, where we varied the number of clusters used in the Gaussian feature extraction process with a step size of 5. The result demonstrates that clustering with a small number will benefit the registration accuracy, while a large clustering number will lead to an intensive drop in performance. The second ablation experiment in Table~\ref{tab:cascad} examined the cascading strategy. We explored different fusion strategies by removing (i) the static features, and (ii) the deformed features. The experimental results indicate that the absence of any of the two features results in a significant degradation in performance. These ablation studies provide valuable insights into the impact of different components in our design. They highlight the importance of selecting an appropriate clustering number and demonstrate the crucial role played by both static and deformed features in achieving accurate and robust surgical scene registration.

\section{Conclusions}

In conclusion, we propose a novel approach for the general implicit 4D dynamic scene registration for preoperative and intraoperative surgery scenes. To target the scene-level dynamic data and, more specifically, surgical scenarios with limited viewpoints, we proposed the Spatially Weighted Clustering and feature cascading strategies, which utilize the spatial relationship of the 4D implicit features and the corresponding coordinate information. This allows both the learning-based and traditional registration methods to outperform the registration of the original point cloud. Furthermore, we present a general registration framework for dynamic surgical scenes. With the proposed method, the registration runs more efficiently due to the size reduction of the features. Experiments on an open surgical dataset demonstrate the effectiveness and superior performance against state-of-the-art methods. Our work paves the way for the real-world applications of surgical robot planning and training.



\bibliographystyle{IEEEtran}
\bibliography{refs}



\end{document}